\documentclass[11pt,a4paper]{article}
\usepackage{acl}

\usepackage{times}
\usepackage{latexsym}

\usepackage[T1]{fontenc}
\usepackage[utf8]{inputenc}

\usepackage{microtype}

\usepackage{amsmath,amssymb}
\usepackage{makecell}
\usepackage{graphicx} 
\usepackage{svg}
\usepackage{subcaption} 
\usepackage{multirow}
\usepackage{color,soul}
\usepackage{enumitem} 
\usepackage{booktabs}
\usepackage{lipsum}
\usepackage{tcolorbox}
\usepackage{amsthm,thmtools}

\declaretheorem[style=definition,name=Definition]{definition}

\declaretheorem[style=remark,numbered=no,name=Remark]{remark}
\declaretheorem[style=remark,numbered=no,name=Proof]{Proof}

\title{On the data requirements of probing}
\author{Zining Zhu$^{1,2}$, Jixuan Wang$^{1,2}$, Bai Li$^{1,2}$, Frank Rudzicz$^{1,2,3}$\\
$^1$ University of Toronto \hspace{0.5em}
$^2$ Vector Institute for Artificial Intelligence \hspace{0.5em} 
$^3$ Unity Health Toronto\\
\texttt{\{zining,jixuan,bai,frank\}@cs.toronto.edu}}

\begin{document}

\maketitle

\begin{abstract}
As large and powerful neural language models are developed, researchers have been increasingly interested in developing diagnostic tools to probe them. There are many papers with conclusions of the form ``observation $X$ is found in model $Y$'', using their own datasets with varying sizes. Larger probing datasets bring more reliability, but are also expensive to collect. There is yet to be a quantitative method for estimating reasonable probing dataset sizes. We tackle this omission in the context of comparing two probing configurations: after we have collected a small dataset from a pilot study, how many additional data samples are sufficient to distinguish two different configurations? We present a novel method to estimate the required number of data samples in such experiments and, across several case studies, we verify that our estimations have sufficient statistical power. Our framework helps to systematically construct probing datasets to diagnose neural NLP models.
\end{abstract}

\section{Introduction}
 While modern deep neural language models achieve impressive performance on various benchmarking datasets, the question of {\em how} this is achieved is gaining increased attention. This line of inquiry includes a new avenue of research: probing.

\begin{figure}[t]
    \centering
    \includegraphics[width=\linewidth]{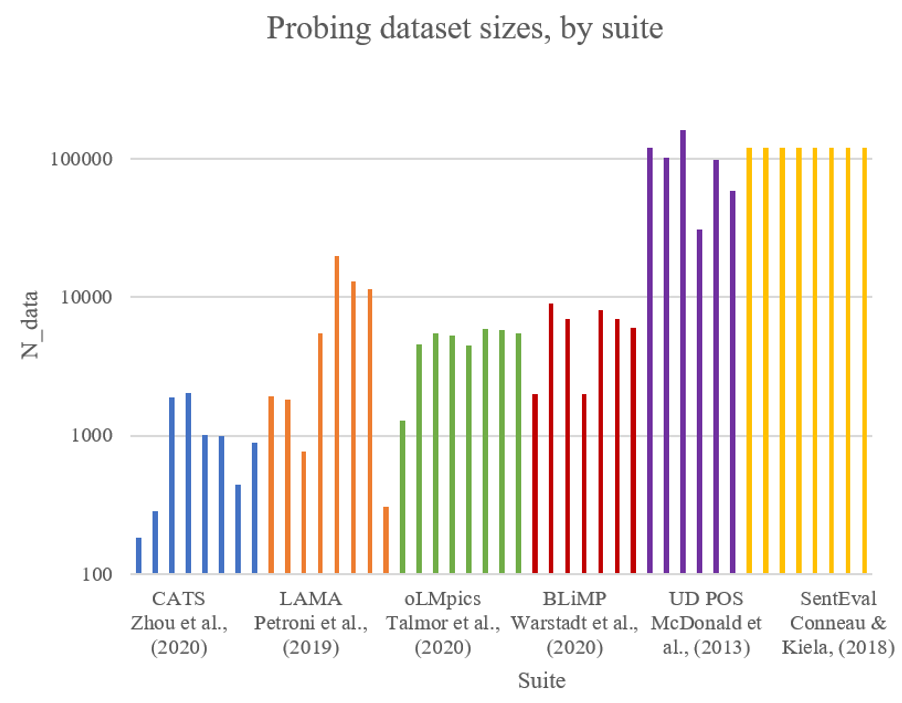}
    \caption{Sizes of the datasets in some common probing suites. Depending on the tasks, they vary from hundreds to larger than $10^5$.}
    \label{fig:probing_dataset_sizes}
\end{figure}

There are many methods to probe a neural network. Among these, diagnostic classification is by far the most common. To probe a neural network in a classification configuration, we specify a classification task that examines an ability (e.g., detecting verb tense). We encode the texts with a deep neural network and apply a post-hoc classifier to the encoded representations. If the classifier can easily predict an attribute, we consider this deep neural network capable of encoding the specified ability.
Researchers have expanded the targets of probing classifications to a wide range of abilities including syntax and semantics \citep{jawahar-etal-2019-bert,tenney-etal-2019-bert,kulmizev-etal-2020-neural,vulic-etal-2020-probing}, discourse \citep{koto-etal-2021-discourse,zhu-etal-2020-examining,chen-etal-2019-evaluation} and commonsense reasoning \citep{petroni-etal-2019-language,lin-etal-2020-birds}. These probing papers are associated with datasets of varying sizes, as shown in Figure \ref{fig:probing_dataset_sizes}. 

What is a suitable size for probing datasets? Larger training datasets lead to tighter generalization bounds. With other conditions fixed, larger testing datasets allow for higher statistical power in comparing probing models and classifiers. That said, it is neither realistic nor desirable to increase the probing dataset sizes arbitrarily. It is therefore essential to find a balanced size for probing classifications.

We propose a framework to formally estimate the data requirements of probing configurations (\S \ref{sec:methodology}). Our framework considers the scenario of comparing probing configurations given some data. How many additional data samples may be necessary to reliably reproduce this comparison effect? We propose a novel method to estimate the required data samples by adapting a generalization bound. 

We evaluate our framework on various probing tasks. First, we verify that the choice of generalization bound agrees with the probing results (\S \ref{subsec:experiment_theoretical_bounds_conservative}). On a case study recognizing synthetic Gaussian noise, we verify that larger datasets provide higher statistical power (\S \ref{subsec:experiment_add_gaussian_noise}). We evaluate the utility of our framework on a collection of comparison problems (\S \ref{subsec:experiment_corrupted_encoders} - \S \ref{subsec:experiment:compare_encoders}), where the probing data sizes recommended by the theoretical framework always supports power larger than $0.8$.
This paper helps to formalize probing experiments. Our framework can be used by the research community in collecting probing datasets.

\section{Related Work}
\paragraph{Understanding the datasets} There is increased research attention paid to the datasets used for deep learning models. 
% Cartography: individual samples
One way to study the dataset is to visualize each of the dataset samples in a map. \citet{swayamdipta-etal-2020-dataset} mapped the data samples in NLP datasets to regions such as ``hard-to-learn'' and ``easy-to-learn'' using signals observed during the training process. \citet{yauney2021comparing} mapped the difficulty of data samples by the data-dependent complexity \citep{arora2019fine}. \citet{vania-etal-2021-comparing} used Item Response Theory \citep{baker2004item}, a statistical framework from psychometrics to describe various attributes related to the difficulty of test set items.
% Datasets as a whole
Similarly, researchers also describe the effects of datasets as a whole. Several papers attempted through information theory \citep{pimentel2021bayesian,zhu_quantifying_2021}. \citet{le-scao-rush-2021-many} compared the effectiveness of prompts to those of classification samples. 
When collecting datasets for probing, showing the effect of data is an important goal. Our framework considers the classification data samples, but we do not impose restrictions on the signal to probe in the classification tasks. 

\paragraph{Probing methods}
The probing literature has proposed numerous diagnostic methods. This paper focuses on diagnostic classifiers, where post-hoc classifiers predict labels from representation vectors. These vectors can be a unified representation of a sequence \citep{conneau_senteval_2018,tenney2019you}, a collection of vectors from different tokens in a sequence \citep{hewitt_structural_2019}, or a pair of carefully set-up vectors that contrast between the ``control'' and the ``treatment'' \citep{hewitt_designing_2019,elazar_amnesic_2021}. We defer to \citet{belinkov_probing_2021} for a systematic overview. 

Note that there are also many probing papers without post-hoc classifiers \citep{zhou-srikumar-2021-directprobe,torroba-hennigen-etal-2020-intrinsic,li-etal-2021-bert}. While many of these do not mention the term ``probing'', they nevertheless probe the intrinsics of deep neural models. In this paper, we consider only classification-based probing methods. Our framework can generalize to unsupervised probing methods in future work.

\paragraph{Reliability of tests}
When all data samples follow i.i.d assumptions, larger datasets allow higher reliability. The reliability of testing is essential in quantitative studies, including medical, societal, and educational contexts \citep{Kraemer1992reliability, Drost2011reliability,golafshani2003understanding}. \textit{Reliability} describes how possible the test results can be reproduced. Depending on the forms of these tests, there are many ways to measure reliability. The test-retest reliability (usually quantified by Pearson correlations between the test and the retest results) measures the consistency of the results across time. The internal consistency reliability (usually quantified by Cronbach's $\alpha$ \citep{cronbach1951coefficient}) measures the consistency of participants responding to a set of items. The inter-rater reliability (quantified by Cohen's $\kappa$ \citep{cohen1960coefficient} or Krippendorff's $\alpha$ \citep{krippendorff2018content}) measures the extent of agreements between the annotators.

To our knowledge, existing works reflecting on the reliability of model diagnostics rely on repeating the tests on a variety of controlled conditions \citep{aribandi-etal-2021-reliable,novikova2021robustness}. A larger variance in results indicates lower test reliability. The reliability is related to two other attributes -- validity and robustness. \textit{Validity} measures how well the test measures what it intends to measure. In a valid test, the result is right for the right reasons \citep{mccoy-etal-2019-right,ravichander-etal-2021-probing}. \textit{Robustness} measures how well the results of a test can generalize from the experimental setting to real-world settings \citep{xing-etal-2020-tasty,niu-etal-2020-evaluating}.

\section{Methodology}
\label{sec:methodology}

\subsection{Problem statement}
\label{subsec:comparison_problem}
Much of probing research considers some form of comparison problem. For example:
\begin{itemize}[nosep]
    \item Which deep neural language model encodes certain linguistic signal in an easier-to-extract manner?
    \item For a neural model, does pretraining with setting $A$ support higher probing classification performances than models pretrained on setting $B$?
    \item Is the accuracy of a simple probing classifier (e.g., logistic regressor) higher than a more complicated one (e.g., MLP with hidden layers)?
\end{itemize}
These problems are instances of comparison problems, where we compare two probing configurations. We formalize them as follows.

\begin{definition}[Probing configuration]
A \textit{probing configuration} $\mathcal{C}$ consists of $\{T, E, f\}$, where $T$ specifies the probing task (e.g., past-vs-present from the SentEval suite), $E$ is the encoder that encodes the text specified by $T$ into representation vectors (e.g., the output from the 11$^{\text{th}}$ layer of BERT\_{base}), and $f$ is the probing classifier.
\label{def:probing_configuration}
\end{definition}
\begin{remark}
Task $T$ can be a text-based classification dataset, with either word inputs or sequential inputs. Here we only consider the classification problems with fixed number of classes. For more complex problems (e.g., generalization problems), the generalization bounds need to be adapted.
\end{remark}

\begin{definition}[Comparison problem]
A \textit{comparison problem} consists of a pair of probing configurations, $\mathcal{C}_A=\{T, E_A, f_A\}$ and $\mathcal{C}_B=\{T, E_B, f_B\}$.
\label{def:comparison_problem}
\end{definition}

\begin{remark}
Usually, the two configurations of a comparison problem, $\mathcal{C}_A$ and $\mathcal{C}_B$ differ by only one of $\{E, f\}$ to avoid confounds. A comparison problem can \textit{collapse} if the two configurations are identical. Recommending the required number of data samples for a collapsed comparison problem is not meaningful.
\end{remark}

\subsection{An overview of the framework}
\label{subsec:method_framework}
For researchers collecting a probing dataset, we recommend the following procedure to estimate the probing dataset sizes:
\begin{enumerate}[nosep]
    \item Identify a comparison problem by specifying two probing configurations, and collect a small set of data in a pilot study. Using the existing data, run the two probing classifications. Let $R_1$ and $R_2$ denote the probing performances, respectively. 
    \item When $|R_1 - R_2|$ is small, it is likely that the comparison problem collapses -- \S \ref{subsec:method_collapsed_comparison_problems} describes some heuristics to verify. 
    \item When the comparison problem has a difference in the performances $|R_1 - R_2|$, our framework can recommend the data requirements: Plug in $\frac{|R_1 - R_2|}{2}$ to the generalization bounds to retrospectively solve for a recommendation of train data size $N_{\text{train}}$. \S \ref{subsec:method_generalization_bounds} elaborates the generalization bounds in probing classifiers. \S \ref{subsec:method_train_data_requirements} presents numerical examples.
    \item Without loss of generality, we assume that the train, validation, and test data have relative sizes of $\eta:1:1$. Then $(1+\frac{2}{\eta})N_{\text{train}}$ is the total data requirement.
\end{enumerate}

\subsection{Generalization bounds}
\label{subsec:method_generalization_bounds}
Machine learning theory literature provides many generalization bounds.
These bounds usually occur in the following form: 
\begin{equation}
\label{eq:generalization_bounds_form}
    \mathbb{P}\left(|R(\hat{f}) - R(f_*)| > B(n,\delta)\right) < \delta,
\end{equation}
where $f$ is the classifier, $R(\cdot)$ is the risk, $n$ is the number of training data points, and $\delta$ is a hyper-parameter. Given $n$, a generalization bound states that, with a probability of at least $1-\delta$, the risk of the empirically optimal classifier $R(\hat{f})$ differs from the risk of the globally optimal classifier $R(f_*)$ by at most $B(n;\delta)$.
The risk is usually assumed to refer to the cross entropy loss. We show that several metrics used in probing have bounds with the form as well.

\paragraph{Accuracy} The most widely used scores to measure the probing performance include accuracy, precision, recall, and F1 score. If we substitute the risk with accuracy, the bounds can apply without loss of generality, modulo two differences: accuracy (etc.) is bounded by $B=1$ (whereas the upper bounds for loss could be larger), and is the highest with $f_*$ (whereas the loss is the lowest with $f_*$).

Note that most behavioural probes\footnote{as described in \citet{belinkov-etal-2020-interpretability}} use evaluation metrics in this category. Many structural probes use additional evaluation metrics. We discuss them below.

\paragraph{Control tasks}
It is possible that the probes, as diagnostic classifiers, rely on some irrelevant dataset statistics to boost the performance. To factor out this effect, \citet{hewitt_structural_2019} proposed to use control tasks.
In a control task setting, we need to set up an auxiliary diagnostic classification task, and take the difference of the two classifications. Note that the difference is related to information theoretic terms \citep{pimentel_information-theoretic_2020,zhu_information_2020}. Regardless, their formulations involve some intractable terms that have to be empirically ignored.

Dependent on the goal of the tasks, there are different ways to set up the auxiliary task. In the part-of-speech (PoS) probing task, for example, \citet{hewitt_structural_2019} associated each word type to a fixed PoS label. Another example is amnesic probing \citep{elazar_amnesic_2021}, which uses iterative null-space projection \citep{ravfogel_null_2020} to remove the probing task information from the representations.

\begin{restatable}[]{theorem}{ThmControlTask}
The probing results for control tasks are subject to the generalization bounds in the following form:
\begin{equation}
\label{eq:generalization_bounds_form_control_task}
    \mathbb{P}\left(|R(\hat{f}) - R(f_*)| > 2B(n,\delta)\right) < \delta
\end{equation}
\end{restatable}

\begin{Proof}
The proofs of all theorems are listed in Appendix \ref{sec:proofs_of_theorems}.
\end{Proof}

\paragraph{Minimum description length}
Recently, \citet{voita_information-theoretic_2020} presented an alternative viewpoint of structural probing based on the minimum description length (MDL). The MDL of a probe or classifier is defined by the sum of (a) the code length required to transmit the data, and (b) the code length required to transmit the model for compressing the data. \citet{voita_information-theoretic_2020} gives two ways to approximate the MDL values: variational and prequential.

The \textit{variational MDL} consists of two terms: the cross entropy loss $L(\hat{f})$, and the KL divergence between the posterior ($\beta$) and prior distribution ($\alpha$) of the model parameters $\theta$. 
$$MDL_{\text{var}} = L(\hat{f}) + KL(\beta_\theta\ \|\ \alpha_\theta)$$

\begin{restatable}[]{theorem}{ThmVariationalMDL}
The probing results of variational MDLs subject to the identical bounds as Eq. \ref{eq:generalization_bounds_form}.
\end{restatable}

The \textit{prequential MDL} computes the code length required in this ``transmission protocol''. First, transmit the first $t_1$ data points using random coding. Then, optimize the model with the transmitted data, and transmit the next portion with the new model. The first portion $t_1$ constitutes of as few as $0.1\%$ of the dataset.
\begin{align*}
MDL_{\text{pre}}&=t_1\text{log}K - \\
&\sum_{i=1}^{S-1}\text{log} p_{f_i} (y_{t_i+1..t_{i+1}}|x_{t_i+1..t_{i+1}}) \\
&=t_1\text{log}K + \sum_{i=1}^{S-1} R(f_i;n_i)
\end{align*}

\begin{restatable}[]{theorem}{ThmPrequentialMDL}
The generalization bound for prequential MDL takes the following form:
\begin{equation}
    \mathbb{P}\left( |R(\hat{f}) - R(f_*)| > \frac{Cn}{t_1}B(n, \delta) \right) < \delta,
\end{equation}
where $C$ is a constant.
\end{restatable}

\subsection{From generalization bounds to training data requirement}
\label{subsec:method_train_data_requirements}
To estimate the required number of training data samples $n$, we can fix $\delta$ and enforce an upper bound on the excess risk $|R(\hat{f}) - R(f_*))|$. Then the corresponding $n$ would be the required number of training data samples. Following is a numerical example where we consider the textbook finite function space bound\footnote{This bound assumes that classifiers $f$ come from a finite space $\mathcal{F}$. We defer to Theorem 7 of \citet{liang2016cs229t} for details. In the rest of this paper, we use this finite function class bound.}: 
$$\mathbb{P}\left(|R(\hat{f}) - R(f_*)| > B\sqrt{\frac{2\text{log}\frac{2|\mathcal{F}|}{\delta}}{n}}\right) < \delta.$$

Here, we set $\delta=10^{-8}$. In a probing classification configuration $\mathcal{C}=\{T, E, f\}$, the encoder $E$ produces vectors with $D=768$ dimensions and $f$ is a logistic regressor. Additionally, we assume that the $D+1$ weight parameters in $f$ are stored in 32-bit floating point numbers\footnote{Following PyTorch's default for \texttt{FloatTensor}.}, so each weight parameter takes $2^{32}$ possible values. Then the probing classifier constitutes a finite space with cardinality $|\mathcal{F}| = 2^{32} \times (D+1)$.

When there are $n=65,536$ training data points, with probability of at least $1-10^{-8}$, the empirically optimal accuracy is different from the global minimum by at most $0.039$ for $D=4,096$ (InferSent) classifications.

More importantly, we can also plug in an expectation on the generalization bound to retrospectively solve for the training data requirement. For example, a bound of $0.05$ requires $N=40\textrm{k}$ i.i.d data samples at $D=4,096$.

If the datasets for both probing configurations in a probing classification are sufficiently large, the generalization bounds would be sufficiently small, so that the result of the comparison problem is reliable. As a heuristic, we let the bound be $\frac{|R_1 - R_2|}{2}$, where $R_1$ and $R_2$ are the probing performances from the existing datasets.

A tighter bound (e.g., $\frac{|R_1 - R_2|}{10}$) requires more data samples (hence larger statistical power) as well as higher expectation for budgets. We consider $\frac{|R_1 - R_2|}{2}$ to be a balanced choice. Following are some justifications.

In the most ideal case, both $R_1$ and $R_2$ are the true global minima $R(f_*)_1$ and $R(f_*)_2$, then the comparison results will remain consistent regardless of the number of data samples.

In the less ideal case, both $R_1$ and $R_2$ are the empirical minima $R(\hat{f})_1$ and $R(\hat{f})_2$, then a generalization bound of $\frac{|R_1 - R_2|}{2}$ guarantees that the relative preference in the comparison will remain consistent (yet the scale of the comparison may fluctuate). We expect that most probing classifiers resemble this scenario, since they reach almost perfect training accuracies.

In the most unfortunate case, $R_1$ and $R_2$ deviate from the empirical minima. The extent they differ contributes to the randomness. While the scales of the empirical imperfectness remain unknown, one can consider some heuristics reduce this imperfectness. First, a probing classifier with higher accuracy tend to have smaller empirical imperfectness, hence smaller unknown instability. Second, identifying the collapsed comparisons helps reduce the uncertainties introduced by the classifier imperfectness. 

\subsection{Power analysis}
\label{subsec:method_power_analysis}
We use power analysis to evaluate the reliability of our data recommendations. For a statistical test, the power is the probability of correctly rejecting the null hypothesis. In the context of this paper, we compare the reliability of the prediction results provided by two probing configurations, $\mathcal{C}_A$ and $\mathcal{C}_B$. The hypothesis is stated as follows.

$H_0$: On a test set $\{x_i\}_{i=1}^{M}$, the results $f_A$ and $f_B$ are not significantly different.

To accept or reject $H_0$ on the two probing classifiers, one can apply the McNemar's test \citep{mcnemar1947note}, which checks if the $\chi^2$ statistic is significant. The $\chi^2$ can be computed as $\chi^2 = \frac{(p_{01} - p_{10})^2}{p_{01} + p_{10}}$, where $p_{00}, p_{01}, p_{10}, p_{11}$ are the probabilities specified by the contingency table (Table \ref{tab:contingency_table}).

\begin{table}[h]
    \centering
    \begin{tabular}{c |c c}
        \toprule
        & $f_B$ incorrect & $f_B$ correct \\ \midrule 
        $f_A$ incorrect & $p_{00}$ & $p_{01}$ \\
        $f_A$ correct & $p_{10}$ & $p_{11}$ \\ \midrule \bottomrule
    \end{tabular}
    \caption{Contingency table between two probing results, $f_A$ and $f_B$.}
    \label{tab:contingency_table}
\end{table}

\citet{card-etal-2020-little} described a framework that estimates the power by simulation. One repetitively samples a portion of test data and computes $\chi^2$. The portion of simulations with significant $\chi^2$ is taken as the estimated power. 
Empirically, one runs multiple classifications with distinct random seeds to increase robustness. To account for multiple classifications, we run the simulations of \citet{card-etal-2020-little} for each random seed, and then count the total number of significant simulations to compute the power.
Usually, we expect that a reliable decision to reject the null hypothesis should have a statistical power of at least $0.8$.

\subsection{Detecting collapsed comparison problems}
\label{subsec:method_collapsed_comparison_problems}
When we have data for a comparison problem from a ``pilot study'' and observe very small classification performance differences (e.g., of $0.5\%$), we might fall back to the null hypothesis -- that the comparison problem collapses -- in this case, increasing the data size does not ``uncollapse'' this comparison problem. Here we describe some heuristics to increase the confidence of detecting a collapse.

In our experiments, our data are subsampled from a larger dataset, so we can test if a probing configuration collapses by repeatedly subsampling the data, and run statistical tests. In the real-world, this is similar to running multiple ``pilot studies'' and collecting small-scale probing data, repeatedly. If the probing configurations output almost indistinguishable results, one can infer that the probing configuration collapses.

Alternatively, one can consider this augmentation method based on cross validation folds. For each dataset in a comparison problem, we divide it into, e.g., 6 folds. For each of $i=1..6$ runs, take Fold $i$ as the validation split, Fold $(i+1)\textrm{ mod }6$ as the test split, and the rest as the train split. Considering the probing classification results of all 6 runs can lead to higher confidence.

\section{Experiments}
\label{sec:experiments}

\subsection{Data and Models}
\label{subsec:data_models}
\paragraph{Probing task}
We run probing classifiers on several classification tasks in one of the largest existing probing suites, SentEval \citep{conneau_senteval_2018}: Past\_present (tense prediction), bigram\_shift (whether two words are flipped in a sentence), and coordination\_inversion (whether two sentences are flipped) are binary classification tasks with 120k samples per class. Sentence\_length contains 6 classes with 12k samples per class. To test the data requirements, we stratify sample subsets with $\{2^7, 2^9, 2^{11}, 2^{13}, 2^{15}\}$ training data samples per class, where applicable\footnote{When there are $2^7$ train sample per class, the subset consists of 256 train samples, 64 dev and 64 test samples, all with evenly distributed labels. i.e., $\eta=4$. In this paper, we keep this ratio consistent.}.

\paragraph{Encoders}
\begin{itemize}[nosep]
    \item BERT \citep{devlin-etal-2019-bert} is a contextualized language model. We take the multilingual $\text{BERT}_{\text{base}}$ model.% and follow the convention to take the 768-dimensional output at [CLS] as the sentence representation vector.
    \item SBERT \citep{reimers-2019-sentence-bert} encouraged semantically similar sentences to be mapped to nearby vectors in the representation space.
    \item GloVe \citep{pennington-etal-2014-glove} is a static word embedding model. It maps each token to a fixed, 300-dimensional vector. We average all embedding vectors of a SentEval sequence as the input representation.
    \item InferSent \citep{infersent} is a contextualized language model. It processes the GloVe embeddings with a bidirectional LSTM \citep{hochreiter1997long} with 2,048 hidden dimensions.
\end{itemize}

\paragraph{Probing classifiers} We use a logistic regressor and a multilayer perceptron (MLP) with 20 hidden units (\S \ref{subsec:experiment_compare_classifiers}) as probing classifiers. In addition, we run several MDL probes, whose results are described in Appendix \ref{subsec:experiment_other_probing_methods}.

\subsection{Verifying the theoretical bounds}
\label{subsec:experiment_theoretical_bounds_conservative}
We run probing classifications using a collection of subsets. Each subset is subsampled in a stratified manner from the dataset. We run 5 probing classifications with different random seeds on each subset.

\begin{figure}[h]
    \centering
    \includesvg[width=\linewidth]{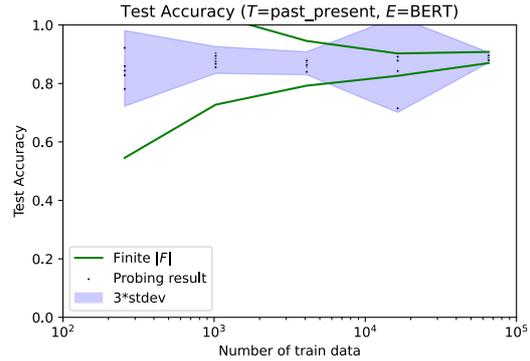}
    %\includesvg[width=.49\linewidth]{figures/past_present_glove_test_acc.svg}
    \caption{Theoretical bounds vs. empirical results on $T=\text{past\_present}$, $E=\text{BERT}$, and $f=\text{LogReg}$. The purple regions represent the empirical margin (mean $\pm$ stdev), while the green lines are the empirical mean $\pm$ margins computed by the learning theory bound.}
    \label{fig:theoretical_vs_empirical}
\end{figure}

To qualitatively examine the extent that the generalization bounds agree with the probing classifications, we plot both the empirical and the theoretical margins. Figure \ref{fig:theoretical_vs_empirical} shows an example. Appendix \ref{subsec:experiment_more_theory_vs_experiment_plots} contains additional plots. The empirical classification results reside within the theoretical margins, except for an outlier classification trial -- this is the classification suboptimality, and we extend the discussion in Appendix \ref{subsec:classifier_suboptimality}.

\subsection{Larger datasets support higher power}
\label{subsec:experiment_add_gaussian_noise}
\begin{figure}[h]
    \centering
    %\includesvg[width=\linewidth]{figures/power_plot_glove.svg}
    \includesvg[width=\linewidth]{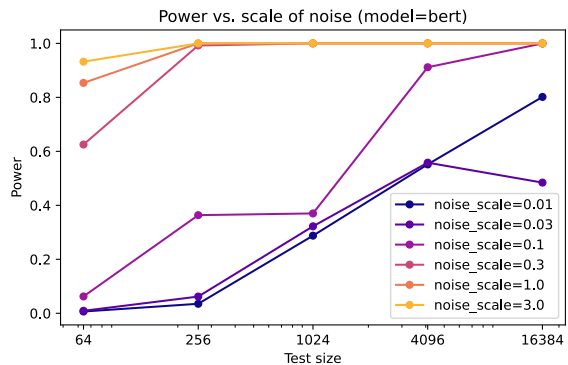}
    \caption{The powers to distinguish the representations with Gaussian noise from the original representations, for $T=\text{past\_present}$, $f=\text{LogReg}$, and $E=\text{BERT}$.}
    \label{fig:power_plot_gaussian}
\end{figure}

Intuitively, adding noise into the representation vectors makes it harder to decode the referential attribute. In this case study, we add Gaussian noise drawn from $\mathcal{N}(0, \sigma^2)$, where $\sigma^2 \in \{.01, .03, .1, .3, 1, 3\}$, and compare against the probing classification with the original representations. Figure \ref{fig:power_plot_gaussian} shows the effect of noise on a configuration. Appendix \ref{subsec:experiment_more_power_vs_noise_plots} contains additional figures. Adding noise with a larger scale results in a configuration that is easier to distinguish. In addition, a larger training dataset usually leads to a higher power to distinguish the configurations.

This case study shows that we can verify the data requirement by incrementally collecting larger datasets for comparisons until we have sufficient power. For example, on the tense prediction task, distinguishing GloVe embeddings from its counterpart with $\mathcal{N}(0, 0.1)$ noise, $1,024$ testing data samples is sufficient to lead to $0.8$ power. However, the same comparison with $\mathcal{N}(0, 0.03)$ noise requires up to $16,384$ testing data points.

The scale of Gaussian noise constitutes a spectrum. When we keep reducing the Gaussian scale, the comparison problem becomes more data-hungry. This leads to a question: where, on the ``min-max'' spectrum, do some other comparison settings (e.g., comparing between encoders) reside? In subsequent case studies, we verify that the numbers predicted by learning theory bounds have sufficient power.

\subsection{Comparing to corrupted encoders}
\label{subsec:experiment_corrupted_encoders}
In this case study, we finetune the BERT models on WikiText\footnote{\texttt{wikitext-2-v1} from huggingface datasets \citep{quentin_lhoest_2021_5510481}.} sentences with scrambled tokens for 200 steps. 

\begin{figure}[h]
    \centering
    \includesvg[width=\linewidth]{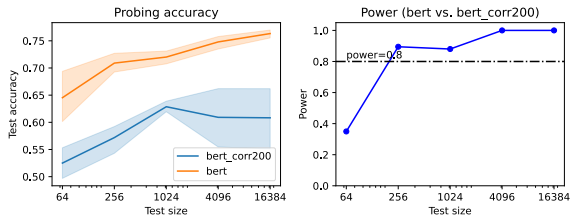}
    \caption{Left: the probing accuracies of BERT (orange) and BERT ``corrupted'' by 200 steps (blue). $T=\text{bigram\_shift}$, $f=\text{LogReg}$. Right: the power to compare between them.}
    \label{fig:comparison_between_encoders_corrupted_bert}
\end{figure}

Table \ref{tab:comparison_between_encoders_corrupted_bert} shows the recommended $N_{\text{train}}$ values in the probing comparisons with corresponding ``pilot data'' (subsampled) sizes. As shown in Figure \ref{fig:comparison_between_encoders_corrupted_bert}, the probing datasets with sizes no fewer than $N_{\text{test}}=256$ (i.e., $N_{\text{train}}=1024$) have sufficient power, and all recommended values fall within the ``sufficient-power'' range.

\begin{table}[h]
    \resizebox{\linewidth}{!}{
    \centering
    \begin{tabular}{r r r}
    \toprule
        Subsampled $N_{\text{test}}$ & Mean $|R_1 - R_2|$ & Recommended $N_{\text{train}}$ \\ \midrule
        64 & .1313 & 22,263\\
        256 & .1281 & 23,362 \\
        1,024 & .0879 & 49,647 \\
        4,096 & .1331 & 21,662 \\
        16,384 & .1488 & 17,331\\ \midrule \bottomrule
    \end{tabular}}
    \caption{The recommended $N_{\text{train}}$ values in the comparison problem in Figure \ref{fig:comparison_between_encoders_corrupted_bert} given different subsample sizes.}
    \label{tab:comparison_between_encoders_corrupted_bert}
\end{table}

\subsection{Comparing between encoders}
\label{subsec:experiment:compare_encoders}
\begin{figure}[h]
    \centering
    \includesvg[width=\linewidth]{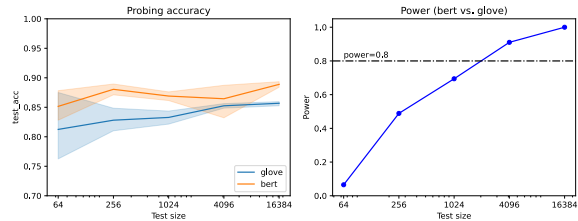}
    \caption{Left: the comparison of accuracies between BERT (orange) and GloVe (blue). $T=\text{past\_present}$, $f=\text{LogReg}$. Right: the power of this comparison. The probing classification accuracy of BERT is higher than that of GloVe, but we do not have enough power to identify that until the testing dataset size is increased to $N_{\text{test}}=4,096$.}
    \label{fig:comparison_between_encoders}
\end{figure}

In this case study, we compare pairs of configurations containing the same task, data, and probing classifier but different encoders. Figure \ref{fig:comparison_between_encoders} shows an example. A test set of size $N_{\text{test}}=1,024$ does not have sufficient power to compare the probing accuracy of BERT vs. GloVe, but $N_{\text{test}}=4,096$ does. This corresponds to $N_{\text{train}=16,384}$, indicating that the recommendations in Table \ref{tab:comparison_between_encoders} are sufficient. Appendix \ref{subsec:experiment_additional_tasks} contains two other examples supporting the same finding.

\begin{table}[h]
    \resizebox{\linewidth}{!}{
    \centering
    \begin{tabular}{r r r}
    \toprule
        Subsampled $N_{\text{test}}$ & Mean $|R_1 - R_2|$ & Recommended $N_{\text{train}}$ \\ \midrule
        64 & .0344 & 324,563 \\
        256 & .0492 & 158,315 \\
        1,024 & .0355 & 303,516 \\
        4,096 & .0091 & 4,600,037 \\
        16,384 & .0320 & 373,513 \\ \midrule \bottomrule
    \end{tabular}}
    \caption{The recommended $N_{\text{train}}$ values in the comparison problem in Figure \ref{fig:comparison_between_encoders} given different subsample sizes.}
    \label{tab:comparison_between_encoders}
\end{table}

\subsection{Comparing between classifiers}
\label{subsec:experiment_compare_classifiers}
\begin{figure}[h]
    \centering
    \includesvg[width=\linewidth]{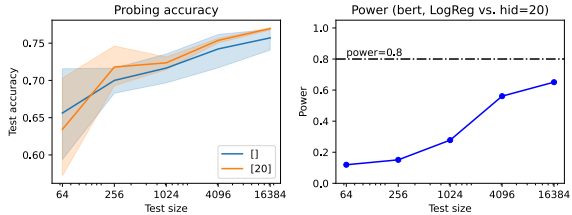}
    \caption{A comparison between probing classifiers. $T=\text{bigram\_shift}$, $E=\text{BERT}$, $f=\{\text{LogReg (blue) vs. MLP (orange)}\}$.}
    \label{fig:comparison_probing_classifiers}
\end{figure}

Here, we compare two probing configurations with different classifiers: LogReg vs. MLP with one hidden layer of $H=20$ neurons. Although the two configurations involve the same task, they have different training data requirements\footnote{The LogReg has $D+1=799$ parameters, while the MLP has $(D+1)H+H+1=15,401$, leading to a larger data requirement.}. We take the larger one as the recommendation.
Table \ref{tab:comparison_probing_classifiers} recommends $N_{\text{train}}$ that are larger than the SentEval dataset sizes. These numbers are actually not necessary --
Figure \ref{fig:comparison_probing_classifiers} shows that $N_{\text{test}}=16.4\textrm{k}$ (corresponding to $N_{\text{train}}=65.6\textrm{k}$) is still insufficient to distinguish the results of the two probing classifier configurations on the bigram\_shift task. There is insufficient evidence to reject the null hypothesis. In other words, the comparison problem between LogReg vs. MLP on $T=\text{bigram\_shift}$ collapses\footnote{Note that LogReg vs. MLP on other tasks e.g., $T=\text{sentence\_length}$ do not collapse. We include the results in Appendix \ref{subsec:experiment_additional_tasks}.}. The exceedingly large data recommendations are meaningless. 

\begin{table}[h]
    \resizebox{\linewidth}{!}{
    \centering
    \begin{tabular}{r r r}
    \toprule
        Subsampled $N_{\text{test}}$ & Mean $|R_1 - R_2|$ & Recommended $N_{\text{train}}$ \\ \midrule
        64 & .022 & 801,472 \\
        256 & .018 & 1,187,812 \\
        1,024 & .007 & 7,757,460 \\
        4,096 & .012 & 2,912,787 \\
        16,384 & .013 & 2,444,949 \\ \midrule \bottomrule
    \end{tabular}}
    \caption{The recommended $N_{\text{train}}$ values in the comparison problem in Figure \ref{fig:comparison_probing_classifiers} given different subsample sizes.}
    \label{tab:comparison_probing_classifiers}
\end{table}

\section{Discussion}
\paragraph{What does a high accuracy entail?}
Our framework implicitly considers the probing classification performances, but the causal relationship between, e.g., the accuracy, the data requirements, and the reliability can be explored further in future frameworks. A high probing accuracy indicates a small empirical risk $R(f)$. This could result from a small $|R(f)-R(\hat{f})|$ (the probe ``learns the task''), or a small $R(\hat{f}) - R(f_*)$ (the distribution of the data samples represent the ``true distribution'' well). The two possibilities resemble the dichotomy raised by \citet{hewitt_designing_2019}, but do they describe the same phenomenon? We leave this as an open question to future researchers.

\paragraph{On the stability of theoretical recommendations}
How stable are our theoretical recommendations? For those comparisons with sufficient evidence to reject the null hypotheses, the recommended $N_{\text{train}}$ sometimes varies (e.g., at $N_{\text{test}}=4096$ in Table \ref{tab:comparison_between_encoders}). This is brought in by the suboptimality of several probing classifications. We extend the discussions about how to interpret and reduce classifier suboptimality in Appendix \ref{subsec:classifier_suboptimality}. Future methods to estimate data requirement may improve the stability.

\paragraph{Cross-task comparison problems} Our framework does not consider cross-task comparison, i.e., when comparing $\mathcal{C}_A=\{T_A, E_A, f_A\}$ vs. $\mathcal{C}_B=\{T_B, E_B, f_B\}$ where $T_A\neq T_B$, because McNemar's test requires pairwise data. Alternative power tests would be necessary to consider cross-task comparison problems. We leave this to an open problem for future research.

\paragraph{Why not just collect as much data as possible?} We argue in favor of knowing how many data samples we need, instead of directly collecting as many samples that budgets allow. The two views resemble the ``top-down'' vs. ``bottom-up'' research approaches mentioned in \citet{bender-koller-2020-climbing}. Practically, our experiments show that many comparison problems do not need as many data samples as the sizes of some existing large probing datasets.

\paragraph{A ``recipe'' for probing datasets} To systematically probe the linguistic abilities of neural networks, many more datasets need to be collected. To make the probing dataset collection procedure systematic, a complete ``recipe'' would be beneficial. Several recent papers called for this goal \citep{ethayarajh-jurafsky-2020-utility,rodriguez-etal-2021-evaluation}. Our framework is one component of such a recipe, by quantifying questions of dataset sizes. Additional components for future work include quantifying the label distributions and the inherent `difficulties' of samples.

\section{Conclusion}
This paper presents a novel framework to estimate the data requirements for probing experiments. This framework uses generalization bounds from formal learning theory to determine minimum training set sizes. In a series of comparison problems, we verify that our recommendations provide sufficient power. Our framework describes an actionable procedure to double check if an experiment needs additional data samples to be scientifically meaningful. Additionally, this paper calls for further attention to complete a systematic methodology in evaluating probing datasets and methods.

\bibliography{bibliography}

\begin{thebibliography}{64}
\expandafter\ifx\csname natexlab\endcsname\relax\def\natexlab#1{#1}\fi

\bibitem[{Aribandi et~al.(2021)Aribandi, Tay, and
  Metzler}]{aribandi-etal-2021-reliable}
Vamsi Aribandi, Yi~Tay, and Donald Metzler. 2021.
\newblock \href {https://doi.org/10.18653/v1/2021.findings-acl.155} {How
  reliable are model diagnostics?}
\newblock In \emph{Findings of the Association for Computational Linguistics:
  ACL-IJCNLP 2021}, pages 1778--1785, Online. Association for Computational
  Linguistics.

\bibitem[{Arora et~al.(2019)Arora, Du, Hu, Li, and Wang}]{arora2019fine}
Sanjeev Arora, Simon Du, Wei Hu, Zhiyuan Li, and Ruosong Wang. 2019.
\newblock Fine-grained analysis of optimization and generalization for
  overparameterized two-layer neural networks.
\newblock In \emph{International Conference on Machine Learning}, pages
  322--332. PMLR.

\bibitem[{Baker and Kim(2004)}]{baker2004item}
Frank~B Baker and Seock-Ho Kim. 2004.
\newblock \emph{Item response theory: Parameter estimation techniques}.
\newblock CRC Press.

\bibitem[{Belinkov(2021)}]{belinkov_probing_2021}
Yonatan Belinkov. 2021.
\newblock \href {http://arxiv.org/abs/2102.12452} {Probing {Classifiers}:
  {Promises}, {Shortcomings}, and {Alternatives}}.
\newblock \emph{arXiv:2102.12452 [cs]}.
\newblock ArXiv: 2102.12452.

\bibitem[{Belinkov et~al.(2020)Belinkov, Gehrmann, and
  Pavlick}]{belinkov-etal-2020-interpretability}
Yonatan Belinkov, Sebastian Gehrmann, and Ellie Pavlick. 2020.
\newblock \href {https://doi.org/10.18653/v1/2020.acl-tutorials.1}
  {Interpretability and analysis in neural {NLP}}.
\newblock In \emph{Proceedings of the 58th Annual Meeting of the Association
  for Computational Linguistics: Tutorial Abstracts}, pages 1--5, Online.
  Association for Computational Linguistics.

\bibitem[{Bender and Koller(2020)}]{bender-koller-2020-climbing}
Emily~M. Bender and Alexander Koller. 2020.
\newblock \href {https://doi.org/10.18653/v1/2020.acl-main.463} {Climbing
  towards {NLU}: {On} meaning, form, and understanding in the age of data}.
\newblock In \emph{Proceedings of the 58th Annual Meeting of the Association
  for Computational Linguistics}, pages 5185--5198, Online. Association for
  Computational Linguistics.

\bibitem[{Card et~al.(2020)Card, Henderson, Khandelwal, Jia, Mahowald, and
  Jurafsky}]{card-etal-2020-little}
Dallas Card, Peter Henderson, Urvashi Khandelwal, Robin Jia, Kyle Mahowald, and
  Dan Jurafsky. 2020.
\newblock \href {https://doi.org/10.18653/v1/2020.emnlp-main.745} {With little
  power comes great responsibility}.
\newblock In \emph{Proceedings of the 2020 Conference on Empirical Methods in
  Natural Language Processing (EMNLP)}, pages 9263--9274, Online. Association
  for Computational Linguistics.

\bibitem[{Chen et~al.(2019)Chen, Chu, and Gimpel}]{chen-etal-2019-evaluation}
Mingda Chen, Zewei Chu, and Kevin Gimpel. 2019.
\newblock \href {https://doi.org/10.18653/v1/D19-1060} {Evaluation benchmarks
  and learning criteria for discourse-aware sentence representations}.
\newblock In \emph{Proceedings of the 2019 Conference on Empirical Methods in
  Natural Language Processing and the 9th International Joint Conference on
  Natural Language Processing (EMNLP-IJCNLP)}, pages 649--662, Hong Kong,
  China. Association for Computational Linguistics.

\bibitem[{Cohen(1960)}]{cohen1960coefficient}
Jacob Cohen. 1960.
\newblock A coefficient of agreement for nominal scales.
\newblock \emph{Educational and psychological measurement}, 20(1):37--46.

\bibitem[{Conneau and Kiela(2018)}]{conneau_senteval_2018}
Alexis Conneau and Douwe Kiela. 2018.
\newblock \href {http://arxiv.org/abs/1803.05449} {{SentEval}: {An}
  {Evaluation} {Toolkit} for {Universal} {Sentence} {Representations}}.
\newblock In \emph{{LREC}}.
\newblock ArXiv: 1803.05449.

\bibitem[{Conneau et~al.(2017)Conneau, Kiela, Schwenk, Barrault, and
  Bordes}]{infersent}
Alexis Conneau, Douwe Kiela, Holger Schwenk, Lo\"{i}c Barrault, and Antoine
  Bordes. 2017.
\newblock \href {https://www.aclweb.org/anthology/D17-1070} {Supervised
  learning of universal sentence representations from natural language
  inference data}.
\newblock In \emph{Proceedings of the 2017 Conference on Empirical Methods in
  Natural Language Processing}, pages 670--680, Copenhagen, Denmark.
  Association for Computational Linguistics.

\bibitem[{Cronbach(1951)}]{cronbach1951coefficient}
Lee~J Cronbach. 1951.
\newblock Coefficient alpha and the internal structure of tests.
\newblock \emph{psychometrika}, 16(3):297--334.

\bibitem[{Devlin et~al.(2019)Devlin, Chang, Lee, and
  Toutanova}]{devlin-etal-2019-bert}
Jacob Devlin, Ming-Wei Chang, Kenton Lee, and Kristina Toutanova. 2019.
\newblock \href {https://doi.org/10.18653/v1/N19-1423} {{BERT}: Pre-training of
  deep bidirectional transformers for language understanding}.
\newblock In \emph{Proceedings of the 2019 Conference of the North {A}merican
  Chapter of the Association for Computational Linguistics: Human Language
  Technologies, Volume 1 (Long and Short Papers)}, pages 4171--4186,
  Minneapolis, Minnesota. Association for Computational Linguistics.

\bibitem[{Drost(2011)}]{Drost2011reliability}
Ellen~A Drost. 2011.
\newblock \href
  {https://search.informit.org/doi/10.3316/informit.491551710186460} {Validity
  and reliability in social science research}.
\newblock \emph{Education Research and Perspectives}, 38(1):105–123.

\bibitem[{Elazar et~al.(2021)Elazar, Ravfogel, Jacovi, and
  Goldberg}]{elazar_amnesic_2021}
Yanai Elazar, Shauli Ravfogel, Alon Jacovi, and Yoav Goldberg. 2021.
\newblock \href {http://arxiv.org/abs/2006.00995} {Amnesic {Probing}:
  {Behavioral} {Explanation} with {Amnesic} {Counterfactuals}}.
\newblock \emph{TACL}.
\newblock ArXiv: 2006.00995.

\bibitem[{Ethayarajh and Jurafsky(2020)}]{ethayarajh-jurafsky-2020-utility}
Kawin Ethayarajh and Dan Jurafsky. 2020.
\newblock \href {https://doi.org/10.18653/v1/2020.emnlp-main.393} {Utility is
  in the eye of the user: A critique of {NLP} leaderboards}.
\newblock In \emph{Proceedings of the 2020 Conference on Empirical Methods in
  Natural Language Processing (EMNLP)}, pages 4846--4853, Online. Association
  for Computational Linguistics.

\bibitem[{Golafshani(2003)}]{golafshani2003understanding}
Nahid Golafshani. 2003.
\newblock Understanding reliability and validity in qualitative research.
\newblock \emph{The qualitative report}, 8(4):597--607.

\bibitem[{Hewitt and Liang(2019)}]{hewitt_designing_2019}
John Hewitt and Percy Liang. 2019.
\newblock \href {https://doi.org/10.18653/v1/D19-1275} {Designing and
  {Interpreting} {Probes} with {Control} {Tasks}}.
\newblock In \emph{Proceedings of the 2019 {Conference} on {Empirical}
  {Methods} in {Natural} {Language} {Processing} and the 9th {International}
  {Joint} {Conference} on {Natural} {Language} {Processing}
  ({EMNLP}-{IJCNLP})}, pages 2733--2743, Hong Kong, China. Association for
  Computational Linguistics.

\bibitem[{Hewitt and Manning(2019)}]{hewitt_structural_2019}
John Hewitt and Christopher~D. Manning. 2019.
\newblock \href {https://doi.org/10.18653/v1/N19-1419} {A {Structural} {Probe}
  for {Finding} {Syntax} in {Word} {Representations}}.
\newblock In \emph{Proceedings of NAACL}, pages 4129--4138, Minneapolis,
  Minnesota. Association for Computational Linguistics.

\bibitem[{Hochreiter and Schmidhuber(1997)}]{hochreiter1997long}
Sepp Hochreiter and J{\"u}rgen Schmidhuber. 1997.
\newblock Long short-term memory.
\newblock \emph{Neural computation}, 9(8):1735--1780.

\bibitem[{Jawahar et~al.(2019)Jawahar, Sagot, and
  Seddah}]{jawahar-etal-2019-bert}
Ganesh Jawahar, Beno{\^\i}t Sagot, and Djam{\'e} Seddah. 2019.
\newblock \href {https://doi.org/10.18653/v1/P19-1356} {What does {BERT} learn
  about the structure of language?}
\newblock In \emph{Proceedings of the 57th Annual Meeting of the Association
  for Computational Linguistics}, pages 3651--3657, Florence, Italy.
  Association for Computational Linguistics.

\bibitem[{Kingma and Ba(2014)}]{kingma2014adam}
Diederik~P Kingma and Jimmy Ba. 2014.
\newblock Adam: A method for stochastic optimization.
\newblock \emph{arXiv preprint arXiv:1412.6980}.

\bibitem[{Koto et~al.(2021)Koto, Lau, and Baldwin}]{koto-etal-2021-discourse}
Fajri Koto, Jey~Han Lau, and Timothy Baldwin. 2021.
\newblock \href {https://doi.org/10.18653/v1/2021.naacl-main.301} {Discourse
  probing of pretrained language models}.
\newblock In \emph{Proceedings of the 2021 Conference of the North American
  Chapter of the Association for Computational Linguistics: Human Language
  Technologies}, pages 3849--3864, Online. Association for Computational
  Linguistics.

\bibitem[{Kraemer(1992)}]{Kraemer1992reliability}
Helena~Chmura Kraemer. 1992.
\newblock \href {https://doi.org/10.1177/096228029200100204} {Measurement of
  reliability for categorical data in medical research}.
\newblock \emph{Statistical Methods in Medical Research}, 1(2):183--199.
\newblock PMID: 1341657.

\bibitem[{Krippendorff(2018)}]{krippendorff2018content}
Klaus Krippendorff. 2018.
\newblock \emph{Content analysis: An introduction to its methodology}.
\newblock Sage publications.

\bibitem[{Kulmizev et~al.(2020)Kulmizev, Ravishankar, Abdou, and
  Nivre}]{kulmizev-etal-2020-neural}
Artur Kulmizev, Vinit Ravishankar, Mostafa Abdou, and Joakim Nivre. 2020.
\newblock \href {https://doi.org/10.18653/v1/2020.acl-main.375} {Do neural
  language models show preferences for syntactic formalisms?}
\newblock In \emph{Proceedings of the 58th Annual Meeting of the Association
  for Computational Linguistics}, pages 4077--4091, Online. Association for
  Computational Linguistics.

\bibitem[{Le~Scao and Rush(2021)}]{le-scao-rush-2021-many}
Teven Le~Scao and Alexander Rush. 2021.
\newblock \href {https://doi.org/10.18653/v1/2021.naacl-main.208} {How many
  data points is a prompt worth?}
\newblock In \emph{Proceedings of the 2021 Conference of the North American
  Chapter of the Association for Computational Linguistics: Human Language
  Technologies}, pages 2627--2636, Online. Association for Computational
  Linguistics.

\bibitem[{Lhoest et~al.(2021)Lhoest, del Moral, von Platen, Wolf, Jernite,
  Thakur, Tunstall, Patil, Drame, Chaumond, Plu, Davison, Brandeis, Scao, Sanh,
  Xu, Patry, McMillan-Major, Schmid, Gugger, Liu, Raw, Lesage, Matussière,
  Debut, Bekman, and Delangue}]{quentin_lhoest_2021_5510481}
Quentin Lhoest, Albert~Villanova del Moral, Patrick von Platen, Thomas Wolf,
  Yacine Jernite, Abhishek Thakur, Lewis Tunstall, Suraj Patil, Mariama Drame,
  Julien Chaumond, Julien Plu, Joe Davison, Simon Brandeis, Teven~Le Scao,
  Victor Sanh, Kevin~Canwen Xu, Nicolas Patry, Angelina McMillan-Major, Philipp
  Schmid, Sylvain Gugger, Steven Liu, Nathan Raw, Sylvain Lesage, Théo
  Matussière, Lysandre Debut, Stas Bekman, and Clément Delangue. 2021.
\newblock \href {https://doi.org/10.5281/zenodo.5510481} {huggingface/datasets:
  1.12.1}.

\bibitem[{Li et~al.(2021)Li, Zhu, Thomas, Xu, and Rudzicz}]{li-etal-2021-bert}
Bai Li, Zining Zhu, Guillaume Thomas, Yang Xu, and Frank Rudzicz. 2021.
\newblock \href {https://doi.org/10.18653/v1/2021.acl-long.325} {How is {BERT}
  surprised? {L}ayerwise detection of linguistic anomalies}.
\newblock In \emph{Proceedings of the 59th Annual Meeting of the Association
  for Computational Linguistics and the 11th International Joint Conference on
  Natural Language Processing (Volume 1: Long Papers)}, pages 4215--4228,
  Online. Association for Computational Linguistics.

\bibitem[{Liang(2016)}]{liang2016cs229t}
Percy Liang. 2016.
\newblock \href {https://web.stanford.edu/class/cs229t/notes.pdf}
  {{CS229T}/{STAT231}: {Statistical} {Learning} {Theory} ({Winter} 2016)}.

\bibitem[{Lin et~al.(2020)Lin, Lee, Khanna, and Ren}]{lin-etal-2020-birds}
Bill~Yuchen Lin, Seyeon Lee, Rahul Khanna, and Xiang Ren. 2020.
\newblock \href {https://doi.org/10.18653/v1/2020.emnlp-main.557} {{B}irds have
  four legs?! {N}umer{S}ense: {P}robing {N}umerical {C}ommonsense {K}nowledge
  of {P}re-{T}rained {L}anguage {M}odels}.
\newblock In \emph{Proceedings of the 2020 Conference on Empirical Methods in
  Natural Language Processing (EMNLP)}, pages 6862--6868, Online. Association
  for Computational Linguistics.

\bibitem[{Lovering et~al.(2021)Lovering, Jha, Linzen, and
  Pavlick}]{lovering_predicting_2021}
Charles Lovering, Rohan Jha, Tal Linzen, and Ellie Pavlick. 2021.
\newblock \href {https://openreview.net/forum?id=mNtmhaDkAr} {Predicting
  {Inductive} {Biases} of {Pre}-{Trained} {Models}}.
\newblock In \emph{{ICLR}}.

\bibitem[{McCoy et~al.(2019)McCoy, Pavlick, and Linzen}]{mccoy-etal-2019-right}
Tom McCoy, Ellie Pavlick, and Tal Linzen. 2019.
\newblock \href {https://doi.org/10.18653/v1/P19-1334} {Right for the wrong
  reasons: Diagnosing syntactic heuristics in natural language inference}.
\newblock In \emph{Proceedings of the 57th Annual Meeting of the Association
  for Computational Linguistics}, pages 3428--3448, Florence, Italy.
  Association for Computational Linguistics.

\bibitem[{McDonald et~al.(2013)McDonald, Nivre, Quirmbach-Brundage, Goldberg,
  Das, Ganchev, Hall, Petrov, Zhang, T{\"a}ckstr{\"o}m
  et~al.}]{mcdonald2013universal}
Ryan McDonald, Joakim Nivre, Yvonne Quirmbach-Brundage, Yoav Goldberg, Dipanjan
  Das, Kuzman Ganchev, Keith Hall, Slav Petrov, Hao Zhang, Oscar
  T{\"a}ckstr{\"o}m, et~al. 2013.
\newblock Universal dependency annotation for multilingual parsing.
\newblock In \emph{Proceedings of the 51st Annual Meeting of the Association
  for Computational Linguistics (Volume 2: Short Papers)}, pages 92--97.

\bibitem[{McNemar(1947)}]{mcnemar1947note}
Quinn McNemar. 1947.
\newblock \href {https://doi.org/10.1007/BF02295996} {Note on the sampling
  error of the difference between correlated proportions or percentages}.
\newblock \emph{Psychometrika}, 12(2):153--157.

\bibitem[{Molchanov et~al.(2017)Molchanov, Ashukha, and
  Vetrov}]{molchanov_variational_2017}
Dmitry Molchanov, Arsenii Ashukha, and Dmitry Vetrov. 2017.
\newblock Variational {Dropout} {Sparsifies} {Deep} {Neural} {Networks}.
\newblock In \emph{Proceedings of {International} {Conference} of {Machine}
  {Learning}}, page~10.

\bibitem[{Niu et~al.(2020)Niu, Mathur, Dinu, and
  Al-Onaizan}]{niu-etal-2020-evaluating}
Xing Niu, Prashant Mathur, Georgiana Dinu, and Yaser Al-Onaizan. 2020.
\newblock \href {https://doi.org/10.18653/v1/2020.acl-main.755} {Evaluating
  robustness to input perturbations for neural machine translation}.
\newblock In \emph{Proceedings of the 58th Annual Meeting of the Association
  for Computational Linguistics}, pages 8538--8544, Online. Association for
  Computational Linguistics.

\bibitem[{Novikova(2021)}]{novikova2021robustness}
Jekaterina Novikova. 2021.
\newblock \href {http://arxiv.org/abs/2109.11888} {{Robustness and Sensitivity
  of BERT Models Predicting Alzheimer's Disease from Text}}.
\newblock \emph{W-NUT @ EMNLP 2021}.

\bibitem[{Pennington et~al.(2014)Pennington, Socher, and
  Manning}]{pennington-etal-2014-glove}
Jeffrey Pennington, Richard Socher, and Christopher Manning. 2014.
\newblock \href {https://doi.org/10.3115/v1/D14-1162} {{G}lo{V}e: Global
  vectors for word representation}.
\newblock In \emph{Proceedings of the 2014 Conference on Empirical Methods in
  Natural Language Processing ({EMNLP})}, pages 1532--1543, Doha, Qatar.
  Association for Computational Linguistics.

\bibitem[{Petroni et~al.(2019)Petroni, Rockt{\"a}schel, Riedel, Lewis, Bakhtin,
  Wu, and Miller}]{petroni-etal-2019-language}
Fabio Petroni, Tim Rockt{\"a}schel, Sebastian Riedel, Patrick Lewis, Anton
  Bakhtin, Yuxiang Wu, and Alexander Miller. 2019.
\newblock \href {https://doi.org/10.18653/v1/D19-1250} {Language models as
  knowledge bases?}
\newblock In \emph{Proceedings of the 2019 Conference on Empirical Methods in
  Natural Language Processing and the 9th International Joint Conference on
  Natural Language Processing (EMNLP-IJCNLP)}, pages 2463--2473, Hong Kong,
  China. Association for Computational Linguistics.

\bibitem[{Pimentel and Cotterell(2021)}]{pimentel2021bayesian}
Tiago Pimentel and Ryan Cotterell. 2021.
\newblock \href {https://arxiv.org/abs/2109.03853} {A {B}ayesian framework for
  information-theoretic probing}.
\newblock \emph{EMNLP}.

\bibitem[{Pimentel et~al.(2020{\natexlab{a}})Pimentel, Saphra, Williams, and
  Cotterell}]{pimentel-etal-2020-pareto}
Tiago Pimentel, Naomi Saphra, Adina Williams, and Ryan Cotterell.
  2020{\natexlab{a}}.
\newblock \href {https://doi.org/10.18653/v1/2020.emnlp-main.254} {{P}areto
  probing: {T}rading off accuracy for complexity}.
\newblock In \emph{Proceedings of the 2020 Conference on Empirical Methods in
  Natural Language Processing (EMNLP)}, pages 3138--3153, Online. Association
  for Computational Linguistics.

\bibitem[{Pimentel et~al.(2020{\natexlab{b}})Pimentel, Valvoda, Hall~Maudslay,
  Zmigrod, Williams, and Cotterell}]{pimentel_information-theoretic_2020}
Tiago Pimentel, Josef Valvoda, Rowan Hall~Maudslay, Ran Zmigrod, Adina
  Williams, and Ryan Cotterell. 2020{\natexlab{b}}.
\newblock \href {https://doi.org/10.18653/v1/2020.acl-main.420}
  {Information-{Theoretic} {Probing} for {Linguistic} {Structure}}.
\newblock In \emph{Proceedings of the 58th {Annual} {Meeting} of the
  {Association} for {Computational} {Linguistics}}, pages 4609--4622, Online.
  Association for Computational Linguistics.

\bibitem[{Ravfogel et~al.(2020)Ravfogel, Elazar, Gonen, Twiton, and
  Goldberg}]{ravfogel_null_2020}
Shauli Ravfogel, Yanai Elazar, Hila Gonen, Michael Twiton, and Yoav Goldberg.
  2020.
\newblock \href {https://doi.org/10.18653/v1/2020.acl-main.647} {Null {It}
  {Out}: {Guarding} {Protected} {Attributes} by {Iterative} {Nullspace}
  {Projection}}.
\newblock In \emph{Proceedings of the 58th {Annual} {Meeting} of the
  {Association} for {Computational} {Linguistics}}, pages 7237--7256, Online.
  Association for Computational Linguistics.

\bibitem[{Ravichander et~al.(2021)Ravichander, Belinkov, and
  Hovy}]{ravichander-etal-2021-probing}
Abhilasha Ravichander, Yonatan Belinkov, and Eduard Hovy. 2021.
\newblock \href {https://aclanthology.org/2021.eacl-main.295} {Probing the
  probing paradigm: Does probing accuracy entail task relevance?}
\newblock In \emph{Proceedings of the 16th Conference of the European Chapter
  of the Association for Computational Linguistics: Main Volume}, pages
  3363--3377, Online. Association for Computational Linguistics.

\bibitem[{Reimers and Gurevych(2019)}]{reimers-2019-sentence-bert}
Nils Reimers and Iryna Gurevych. 2019.
\newblock \href {http://arxiv.org/abs/1908.10084} {{Sentence-BERT: Sentence
  Embeddings using Siamese BERT-Networks}}.
\newblock In \emph{Proceedings of the 2019 Conference on Empirical Methods in
  Natural Language Processing}. Association for Computational Linguistics.

\bibitem[{Ribeiro et~al.(2020)Ribeiro, Wu, Guestrin, and
  Singh}]{ribeiro_beyond_2020}
Marco~Tulio Ribeiro, Tongshuang Wu, Carlos Guestrin, and Sameer Singh. 2020.
\newblock \href {https://doi.org/10.18653/v1/2020.acl-main.442} {Beyond
  {Accuracy}: {Behavioral} {Testing} of {NLP} {Models} with {CheckList}}.
\newblock In \emph{Proceedings of the 58th {Annual} {Meeting} of the
  {Association} for {Computational} {Linguistics}}, pages 4902--4912, Online.
  Association for Computational Linguistics.

\bibitem[{Rodriguez et~al.(2021)Rodriguez, Barrow, Hoyle, Lalor, Jia, and
  Boyd-Graber}]{rodriguez-etal-2021-evaluation}
Pedro Rodriguez, Joe Barrow, Alexander~Miserlis Hoyle, John~P. Lalor, Robin
  Jia, and Jordan Boyd-Graber. 2021.
\newblock \href {https://doi.org/10.18653/v1/2021.acl-long.346} {Evaluation
  examples are not equally informative: How should that change {NLP}
  leaderboards?}
\newblock In \emph{Proceedings of the 59th Annual Meeting of the Association
  for Computational Linguistics and the 11th International Joint Conference on
  Natural Language Processing (Volume 1: Long Papers)}, pages 4486--4503,
  Online. Association for Computational Linguistics.

\bibitem[{Swayamdipta et~al.(2020)Swayamdipta, Schwartz, Lourie, Wang,
  Hajishirzi, Smith, and Choi}]{swayamdipta-etal-2020-dataset}
Swabha Swayamdipta, Roy Schwartz, Nicholas Lourie, Yizhong Wang, Hannaneh
  Hajishirzi, Noah~A. Smith, and Yejin Choi. 2020.
\newblock \href {https://doi.org/10.18653/v1/2020.emnlp-main.746} {Dataset
  cartography: Mapping and diagnosing datasets with training dynamics}.
\newblock In \emph{Proceedings of the 2020 Conference on Empirical Methods in
  Natural Language Processing (EMNLP)}, pages 9275--9293, Online. Association
  for Computational Linguistics.

\bibitem[{Talmor et~al.(2020)Talmor, Elazar, Goldberg, and
  Berant}]{talmor-etal-2020-olmpics}
Alon Talmor, Yanai Elazar, Yoav Goldberg, and Jonathan Berant. 2020.
\newblock \href {https://doi.org/10.1162/tacl_a_00342} {o{LM}pics-on what
  language model pre-training captures}.
\newblock \emph{Transactions of the Association for Computational Linguistics},
  8:743--758.

\bibitem[{Tenney et~al.(2019{\natexlab{a}})Tenney, Das, and
  Pavlick}]{tenney-etal-2019-bert}
Ian Tenney, Dipanjan Das, and Ellie Pavlick. 2019{\natexlab{a}}.
\newblock \href {https://doi.org/10.18653/v1/P19-1452} {{BERT} rediscovers the
  classical {NLP} pipeline}.
\newblock In \emph{Proceedings of the 57th Annual Meeting of the Association
  for Computational Linguistics}, pages 4593--4601, Florence, Italy.
  Association for Computational Linguistics.

\bibitem[{Tenney et~al.(2019{\natexlab{b}})Tenney, Xia, Chen, Wang, Poliak,
  McCoy, Kim, Van~Durme, Bowman, Das et~al.}]{tenney2019you}
Ian Tenney, Patrick Xia, Berlin Chen, Alex Wang, Adam Poliak, R~Thomas McCoy,
  Najoung Kim, Benjamin Van~Durme, Samuel~R Bowman, Dipanjan Das, et~al.
  2019{\natexlab{b}}.
\newblock {What do you learn from context? {P}robing for sentence structure in
  contextualized word representations}.
\newblock In \emph{ICLR}.

\bibitem[{Torroba~Hennigen et~al.(2020)Torroba~Hennigen, Williams, and
  Cotterell}]{torroba-hennigen-etal-2020-intrinsic}
Lucas Torroba~Hennigen, Adina Williams, and Ryan Cotterell. 2020.
\newblock \href {https://doi.org/10.18653/v1/2020.emnlp-main.15} {Intrinsic
  probing through dimension selection}.
\newblock In \emph{Proceedings of the 2020 Conference on Empirical Methods in
  Natural Language Processing (EMNLP)}, pages 197--216, Online. Association for
  Computational Linguistics.

\bibitem[{Vania et~al.(2021)Vania, Htut, Huang, Mungra, Pang, Phang, Liu, Cho,
  and Bowman}]{vania-etal-2021-comparing}
Clara Vania, Phu~Mon Htut, William Huang, Dhara Mungra, Richard~Yuanzhe Pang,
  Jason Phang, Haokun Liu, Kyunghyun Cho, and Samuel~R. Bowman. 2021.
\newblock \href {https://doi.org/10.18653/v1/2021.acl-long.92} {Comparing test
  sets with item response theory}.
\newblock In \emph{Proceedings of the 59th Annual Meeting of the Association
  for Computational Linguistics and the 11th International Joint Conference on
  Natural Language Processing (Volume 1: Long Papers)}, pages 1141--1158,
  Online. Association for Computational Linguistics.

\bibitem[{Voita and Titov(2020)}]{voita_information-theoretic_2020}
Elena Voita and Ivan Titov. 2020.
\newblock \href {https://doi.org/10.18653/v1/2020.emnlp-main.14}
  {Information-{Theoretic} {Probing} with {Minimum} {Description} {Length}}.
\newblock In \emph{Proceedings of the 2020 {Conference} on {Empirical}
  {Methods} in {Natural} {Language} {Processing} ({EMNLP})}, pages 183--196,
  Online. Association for Computational Linguistics.

\bibitem[{Vuli{\'c} et~al.(2020)Vuli{\'c}, Ponti, Litschko, Glava{\v{s}}, and
  Korhonen}]{vulic-etal-2020-probing}
Ivan Vuli{\'c}, Edoardo~Maria Ponti, Robert Litschko, Goran Glava{\v{s}}, and
  Anna Korhonen. 2020.
\newblock \href {https://doi.org/10.18653/v1/2020.emnlp-main.586} {Probing
  pretrained language models for lexical semantics}.
\newblock In \emph{Proceedings of the 2020 Conference on Empirical Methods in
  Natural Language Processing (EMNLP)}, pages 7222--7240, Online. Association
  for Computational Linguistics.

\bibitem[{Warstadt et~al.(2020)Warstadt, Parrish, Liu, Mohananey, Peng, Wang,
  and Bowman}]{warstadt-etal-2020-blimp-benchmark}
Alex Warstadt, Alicia Parrish, Haokun Liu, Anhad Mohananey, Wei Peng, Sheng-Fu
  Wang, and Samuel~R. Bowman. 2020.
\newblock \href {https://doi.org/10.1162/tacl_a_00321} {{BL}i{MP}: The
  benchmark of linguistic minimal pairs for {E}nglish}.
\newblock \emph{Transactions of the Association for Computational Linguistics},
  8:377--392.

\bibitem[{Xing et~al.(2020)Xing, Jin, Jin, Wang, Zhang, and
  Huang}]{xing-etal-2020-tasty}
Xiaoyu Xing, Zhijing Jin, Di~Jin, Bingning Wang, Qi~Zhang, and Xuanjing Huang.
  2020.
\newblock \href {https://doi.org/10.18653/v1/2020.emnlp-main.292} {Tasty
  burgers, soggy fries: Probing aspect robustness in aspect-based sentiment
  analysis}.
\newblock In \emph{Proceedings of the 2020 Conference on Empirical Methods in
  Natural Language Processing (EMNLP)}, pages 3594--3605, Online. Association
  for Computational Linguistics.

\bibitem[{Yauney and Mimno(2021)}]{yauney2021comparing}
Gregory Yauney and David Mimno. 2021.
\newblock Comparing text representations: A theory-driven approach.
\newblock \emph{arXiv preprint arXiv:2109.07458}.

\bibitem[{Zhou et~al.(2020)Zhou, Zhang, Cui, and Huang}]{zhou2020evaluating}
Xuhui Zhou, Yue Zhang, Leyang Cui, and Dandan Huang. 2020.
\newblock Evaluating commonsense in pre-trained language models.
\newblock In \emph{Proceedings of the AAAI Conference on Artificial
  Intelligence}, volume~34, pages 9733--9740.

\bibitem[{Zhou and Srikumar(2021)}]{zhou-srikumar-2021-directprobe}
Yichu Zhou and Vivek Srikumar. 2021.
\newblock \href {https://doi.org/10.18653/v1/2021.naacl-main.401}
  {{D}irect{P}robe: Studying representations without classifiers}.
\newblock In \emph{Proceedings of the 2021 Conference of the North American
  Chapter of the Association for Computational Linguistics: Human Language
  Technologies}, pages 5070--5083, Online. Association for Computational
  Linguistics.

\bibitem[{Zhu et~al.(2021)Zhu, Balagopalan, Ghassemi, and
  Rudzicz}]{zhu_quantifying_2021}
Zining Zhu, Aparna Balagopalan, Marzyeh Ghassemi, and Frank Rudzicz. 2021.
\newblock \href {https://arxiv.org/abs/2110.08931} {Quantifying the
  {Task}-{Specific} {Information} in {Text}-{Based} {Classifications}}.
\newblock \emph{arXiv preprint arXiv:2110.08931}.

\bibitem[{Zhu et~al.(2020)Zhu, Pan, Abdalla, and
  Rudzicz}]{zhu-etal-2020-examining}
Zining Zhu, Chuer Pan, Mohamed Abdalla, and Frank Rudzicz. 2020.
\newblock \href {https://doi.org/10.18653/v1/2020.blackboxnlp-1.3} {Examining
  the rhetorical capacities of neural language models}.
\newblock In \emph{Proceedings of the Third BlackboxNLP Workshop on Analyzing
  and Interpreting Neural Networks for NLP}, pages 16--32, Online. Association
  for Computational Linguistics.

\bibitem[{Zhu and Rudzicz(2020)}]{zhu_information_2020}
Zining Zhu and Frank Rudzicz. 2020.
\newblock \href {https://www.aclweb.org/anthology/2020.emnlp-main.744/} {An
  information theoretic view on selecting linguistic probes}.
\newblock In \emph{{EMNLP}}, pages 9251--9262. Association for Computational
  Linguistics.

\end{thebibliography}
\bibliographystyle{acl_natbib}

\newpage 
.
\newpage 
\appendix 

\section{Probing is a unique classification problem}
The learning theory literature provides a rich collection of theories for classification. One might consider that these theories can directly apply to probing classifiers, but we argue for the alternative. Compared to conventional classifiers, the probing classifiers differ in many aspects. Following are some of them.

\textit{Goals.} Conventional classifiers try to reach high performance on both the experimental and the real-world data distribution. Probing classifiers, while also maximize the probing performance, aim at quantifying the ``easiness to decode'' from the input representations \citep{belinkov_probing_2021}. Therefore, some papers (e.g., \citet{hewitt_designing_2019}) argued in favor of selectivity. A highly selective probe should output results that differ a lot between hard-to-decode and easy-to-decode representations.

\textit{Models.} In general, conventional classifiers use models with many more parameters than probing classifiers. Researchers have raised several concerns for larger probing classifiers. First, larger probing classifiers may ``learn the task'', introducing a confounding factor in the result interpretation: a high probing classification performance could result from the probe itself ``learns the task''. \citet{hewitt_designing_2019} raised this hypothesis, and \citet{zhu_information_2020} confirmed from an information-theoretic perspective. Second, larger probing classifiers require more data to train, which slows down the diagnosis procedure. Ideally, the computation effort spent in diagnosis should be much smaller than training the neural models.

\textit{Datasets.} The data to train a conventional classifier should be abundant, so the classifier could learn sufficient inductive bias that can generalize beyond the experimental conditions. The data to train a probing classifier, however, should contain a collection of specific ``test cases'', covering the ``corner cases'' of the deep neural models, akin to the diagnostic suites in software engineering \citep{ribeiro_beyond_2020}. 

Considering the differences, it is necessary to formulate a framework to study the validity of probing rigorously. Adapting the tools in machine learning theory can be a good start.

\section{Proofs of theorems}
\label{sec:proofs_of_theorems}

\ThmControlTask*
\begin{Proof}
Let us use $A_o$ and $A_c$ to denote the original and the control task performance, and $\hat{f}$ and $f_*$ (and $\hat{f}_c$ and $\hat{f}_{c*}$ for control task) to denote the empirical and the optimal classifier, respectively.

Since both $A_o$ and $A_c$ are count-based metrics, the aforementioned analysis gives us $|A_o(\hat{f}) - A_o(f_*)|\leq B_o$, and $|A_c(\hat{f}_c) - A_c(f_{c*})| \leq B_c$ with probabilities $1-\delta_o$ and $1-\delta_c$ respectively.

Then, with probability $(1-\delta_o)(1-\delta_c)$, we have $|A_o(\hat{f}) - A_c(\hat{f}_c) - \left(A_o(f_*) - A_c(f_{c*}) \right)| \leq B_o + B_c$. In other words, a bound with the same form, hence the same convergence rate, as the count-based metrics still applies to the results of control tasks.
\qed
\end{Proof}

\ThmVariationalMDL*
\begin{Proof} 
The generalization error of variational MDL is bounded by that of $R(\cdot)$, plus the estimation uncertainty of $\text{KL}(\beta_\theta\ \|\ \alpha_\theta)$. In a Bayesian network implementation, the $\text{KL}(\beta_\theta\ \|\ \alpha_\theta)$ can be acquired with less than 2e-3 variance \citep{molchanov_variational_2017}, bringing in a negligible additional uncertainty. In short, the generalization of variational MDL is bounded by the cross entropy term.
\qed 
\end{Proof}

\ThmPrequentialMDL*
\begin{Proof} 
Following likewise analysis, the generalization error of individual loss term is bounded by $\epsilon(n_i) = R(f_i;n_i) - R(f_{i*}) \leq B(n_i, \delta)$ with probability of at least $1-\delta$.

Using a union bound, the error of summing up all these terms is bounded by the sum of all individual bound. The error bound of prequential MDL is dominated by the first a few terms (i.e., the cross entropy losses with $n=\{0.1\%, 0.2\%, ... \}N$). 
\qed 
\end{Proof}

\begin{remark}
Naturally, the theoretical bounds for prequential MDL appear ``looser'' than the bounds of previous metrics.
\end{remark}

\begin{figure*}
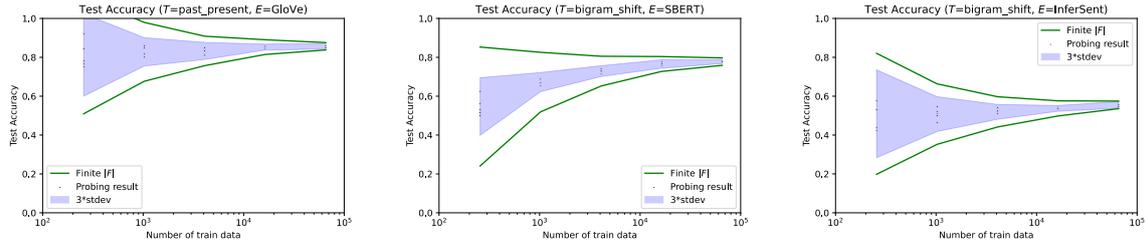

    \centering
    \includesvg[width=.32\linewidth]{figures/past_present_glove_test_acc.svg}
    \includesvg[width=.32\linewidth]{figures/bigram_shift_sbert_test_acc.svg}
    \includesvg[width=.32\linewidth]{figures/bigram_shift_infersent_test_acc.svg}
    \caption{Theoretical bounds vs. empirical results, with $f=\text{LogReg}$. Left: $T=\text{past\_present}$, $E=\text{GloVe}$. Middle: $T=\text{bigram\_shift}$, $E=\text{SBERT}$. Right: $T=\text{bigram\_shift}$, $E=\text{InferSent}$.}
    \label{fig:theoretical_vs_empirical_additional}
\end{figure*}

\section{Additional experiment details}
\subsection{Theory bounds vs experiment plots}
\label{subsec:experiment_more_theory_vs_experiment_plots}
We include additional theory vs. experiment plots in Figure \ref{fig:theoretical_vs_empirical_additional}. The purple regions represent the empirical margin (mean $\pm$ stdev), while the lines represent the margins predicted by the empirical mean $\pm$ the learning theory bound.

\begin{figure*}
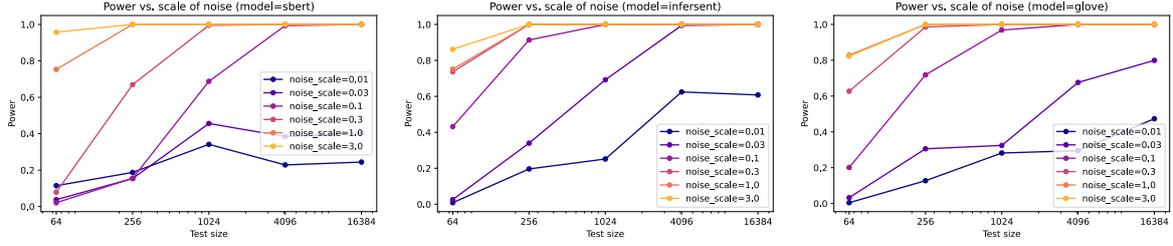

    \centering 
    \includesvg[width=.32\linewidth]{figures/power_plot_sbert.svg}
    \includesvg[width=.32\linewidth]{figures/power_plot_infersent.svg}
    \includesvg[width=.32\linewidth]{figures/power_plot_glove.svg}
    \caption{Additional power vs. scale of noise plots, on $T=\text{past\_present}$, $f=\text{LogReg}$. Left: $E=\text{SBERT}$. Middle: $E=\text{InferSent}$. Right: $E=\text{GloVe}$.}
    \label{fig:power_vs_noise_scale_plots_additional}
\end{figure*}

\subsection{Power vs scale of noise plots}
\label{subsec:experiment_more_power_vs_noise_plots}
We include additional power vs. scale of noise plots in Figure \ref{fig:power_vs_noise_scale_plots_additional}.

\subsection{Results on additional tasks}
\label{subsec:experiment_additional_tasks}
\begin{figure}[h]
    \centering
    \includesvg[width=\linewidth]{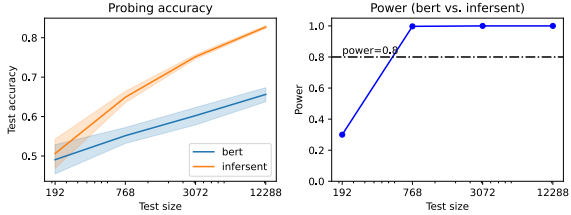}
    \caption{An example for $T=\text{sentence\_length}$. Left: the probing performance of BERT (blue) and InferSent (orange). Right: the statistical power in this comparison.}
    \label{fig:comparison_sentlen_bert_infersent}
\end{figure}

\begin{figure}[h]
    \centering
    \includesvg[width=\linewidth]{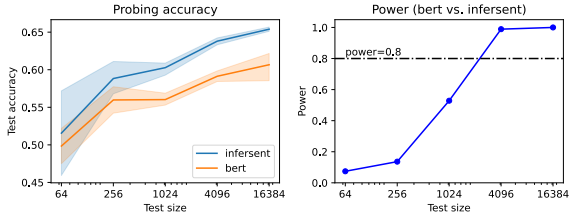}
    \caption{An example for $T=\text{coordination\_inversion}$. Left: the probing performance of BERT (orange) vs. InferSent (blue). Right: the statistical power in this comparison.}
    \label{fig:comparison_coordinv_bert_infersent}
\end{figure}

\begin{figure}[h]
    \centering
    \includesvg[width=\linewidth]{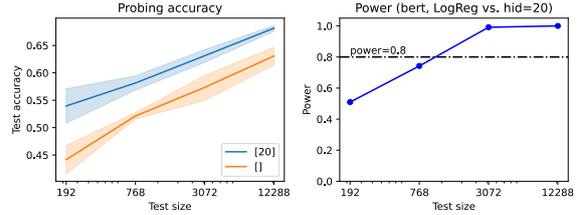}
    \caption{An example for $T=\text{sentence\_length}$. Left: the probing performance of $f$=LogReg (orange) vs. $f$=MLP (blue). Right: the statistical power in this comparison.}
    \label{fig:comparison_sentlen_logreg_mlp}
\end{figure}

We list some results of additional tasks here:
\begin{itemize}[nosep]
    \item Table \ref{tab:comparison_sentlen_bert_infersent} and Figure \ref{fig:comparison_sentlen_bert_infersent} for $T$=sentence\_length, BERT vs InferSent, $f$=LogReg.
    \item Table \ref{tab:comparison_coordinv_bert_infersent} and Figure \ref{fig:comparison_coordinv_bert_infersent} for $T$=coordination\_ inversion, BERT vs InferSent, $f$=LogReg.
    \item Table \ref{tab:comparison_sentlen_logreg_mlp} and Figure \ref{fig:comparison_sentlen_logreg_mlp} for $T$=sentence\_length, $E$=BERT, LogReg vs MLP.
\end{itemize}

% $T$=sentence\_length, BERT vs InferSent, $f$=LogReg

\begin{table}[h]
    \resizebox{\linewidth}{!}{
    \centering
    \begin{tabular}{r r r}
    \toprule
        Subsampled $N_{\text{test}}$ & Mean $|R_1 - R_2|$ & Recommended $N_{\text{train}}$ \\ \midrule
        192 & 0.065 & 95,156 \\
        768 & 0.128 & 24,375 \\
        3,072 & 0.178 & 12,481 \\
        12,288 & 0.196 & 10,285 \\ \midrule \bottomrule
    \end{tabular}}
    \caption{The recommended $N_{\text{train}}$ values in the comparison problem in Figure \ref{fig:comparison_sentlen_bert_infersent}. Given different subsample sizes, the recommended $N_{\text{train}}$ are greater than $3,072$ (i.e., $N_{\text{test}}>768$). Their statistical powers are greater than $0.8$.}
    \label{tab:comparison_sentlen_bert_infersent}
\end{table}

% $T$=coordination\_inversion, BERT vs InferSent, $f$=LogReg
\begin{table}[h]
    \resizebox{\linewidth}{!}{
    \centering
    \begin{tabular}{r r r}
    \toprule 
        Subsampled $N_{\text{test}}$ & Mean $|R_1 - R_2|$ & Recommended $N_{\text{train}}$ \\ \midrule
        64 & 0.025 & 635,040 \\
        256 & 0.019 & 1,128,961 \\
        1,024 & 0.041 & 231,499 \\
        4,096 & 0.051 & 151,861 \\
        16,384 & 0.063 & 100,153 \\ \midrule \bottomrule
    \end{tabular}}
    \caption{The recommended $N_{\text{train}}$ values in the comparison problem of Figure \ref{fig:comparison_coordinv_bert_infersent}. These values correspond to $N_{\text{train}}>4096$, indicating statistical powers of greater than $0.8$.}
    \label{tab:comparison_coordinv_bert_infersent}
\end{table}

% $T$=sentence\_length, BERT, LogReg vs MLP
\begin{table}[h]
    \resizebox{\linewidth}{!}{
    \centering
    \begin{tabular}{r r r}
    \toprule 
        Subsampled $N_{\text{test}}$ & Mean $|R_1 - R_2|$ & Recommended $N_{\text{train}}$ \\ \midrule
        192 & 0.049 & 40,001 \\
        768 & 0.030 & 105,979 \\
        3,072 & 0.029 & 114,746 \\
        12,288 & 0.025 & 148,437 \\ \midrule \bottomrule
    \end{tabular}}
    \caption{The recommended $N_{\text{train}}$ values in the comparison problem of Figure \ref{fig:comparison_sentlen_logreg_mlp}. The accuracies from MLP are higher than that of LogReg, but we do not have sufficient power until $N_{test}=3,072$. This corresponds to $N_{train}=12,288$, which the recommended $N_{train}$ satisfy.}
    \label{tab:comparison_sentlen_logreg_mlp}
\end{table}

\subsection{Hyperparameter configurations}
We use Ray Tune to find the optimal hyperparameters for training. The search space include:\begin{itemize}[nosep]
    \item Learning rate: 1e-4, 5e-4, 1e-3, 5e-3, 1e-2
    \item Batch size: 8, 16, 32, 64
    \item Number of epochs: we set it to 50. We stop running when the validation loss reaches a plateau for 5 epochs. Then, we report the result from the epoch with the highest validation accuracy.
\end{itemize}
We use pytorch to implement the models, and Adam \citep{kingma2014adam} to optimize. To reduce the training time, we cache the representation vectors. The runtime is about one minute per 200 training data points. Our analysis scripts are available at \url{https://github.com/SPOClab-ca/probing_dataset}.

\subsection{Other probing methods}
\label{subsec:experiment_other_probing_methods}
To show the generalizability of our framework, we extend the experiments to two probing classifiers motivated by minimum description lengths: variational and prequential MDL probes \citep{voita_information-theoretic_2020}.
For the variational MDL probe, its results are affected by the arbitrary choice of prior \citep{pimentel-etal-2020-pareto}. Empirically, when we apply a uniform prior, the variational MDL usually degenerates\footnote{The classifiers output the same label for all inputs.}, resulting in 0.5 accuracy. To alleviate this problem, varying of hidden layers and neurons in the probing classifiers is beneficial. For the prequential MDL probes, the results depend on the input sequence of data. \citet{lovering_predicting_2021} mentioned that the early steps sometimes produce cross-entropy losses that are larger than the uniform coding codelength. We also observe this effect, especially when the early steps contain imbalanced data.

\section{Additional discussions}
\subsection{The suboptimality of probing classifier results} 
\label{subsec:classifier_suboptimality}
\begin{figure}[h]
    \centering
    \includegraphics[width=.8\linewidth]{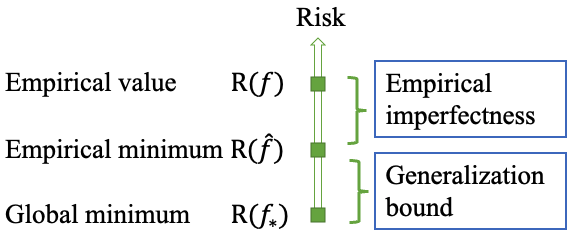}
    \caption{An illustration of the risk values.}
    \label{fig:risk_values}
\end{figure}

The probing classifiers are usually imperfect. Due to the presence of, e.g., degenerative runs and local minima, the empirical result $R(f)$ may be different from the empirical optimum $R(\hat{f})$. While $R(f_*)$ describes the probing classification goal, the ``easiness to extract'', only $R(f)$ is empirically visible. As illustrated in Figure \ref{fig:risk_values}, the difference between the measured values $R(f)$ and the true global minimum $R(f_*)$ can be decomposed into two parts: $R(\hat{f}) - R(f_*)$, which is bounded by the generalization bounds, and $R(f) - R(\hat{f})$, which is the the empirical imperfectness. 

\section{Current sizes of some probing datasets}
This section surveys some commonly used probing datasets, as well as their sizes. Table \ref{tab:probing_dataset_sizes_fixed_class} lists the number of classes and the total number of samples. As listed in the table, most probing classification tasks contain more than enough data for, e.g., comparing BERT vs. InferSent. Regardless, we recommend that future researchers to consider the data requirements for reliability when collecting probing datasets. 

Here is how we count the numbers of data samples: For SentEval \citep{conneau_senteval_2018}, each data sample contains a text sequence and a label. Universal Dependencies \citep{mcdonald2013universal} contains rich annotations, and have been used as, e.g., the part-of-speech tagging probing task \citep{pimentel_information-theoretic_2020}. Here we count the number of words in the train, validation, and the test set respectively. For BLiMP \citep{warstadt-etal-2020-blimp-benchmark}, there are multiple ``phenomenon'' categories for each task, with 1,000 pairs in each phenomenon. The oLMpics \citep{talmor-etal-2020-olmpics} suite splits the task datasets train and test divisions, and we list the numbers of both.

\begin{table}[t]
    \centering
    \resizebox{\linewidth}{!}{
    \begin{tabular}{l r l}
    \toprule 
        Task & N. classes & N. samples \\ \midrule 
        \multicolumn{3}{l}{SentEval \citep{conneau_senteval_2018}} \\
        word\_content & 1,000 & 120k \\
        top\_constituents & 20 & 120k \\
        tree\_depth & 7 & 120k \\
        sentence\_length & 6 & 120k \\
        \makecell{past\_present, bigram\_shift,\\coord\_inv, obj\_num} & 2 & 120k each \\ \midrule
        \multicolumn{3}{l}{UD part-of-speech \citep{mcdonald2013universal}} \\
        Basque & 16 & 73k / 24k / 24k \\
        English & 17 & 70k / 16k / 16k \\
        Finnish & 18 & 128k / 16k / 16k \\
        Marathi & 16 & 3k / 479 / 448\\
        Russian & 16 & 75k / 12k / 11k \\
        Turkish & 15 & 39k / 10k / 10k \\ \midrule
        \multicolumn{3}{l}{BLiMP (selected) \citep{warstadt-etal-2020-blimp-benchmark}} \\
        anaphor\_agreement & 2 & 2k \\
        argument\_structure & 2 & 9k \\ 
        binding & 2 & 7k \\
        ellipsis & 2 & 2k \\
        island effects & 2 & 8k \\
        NPI licensing & 2 & 7k \\
        subject-verb agreement & 2 & 6k \\ \midrule
        \multicolumn{3}{l}{oLMpics \citep{talmor-etal-2020-olmpics}} \\
        Always-Never & 5 & 1,004 / 280 \\
        Age-Comparison & 2 & 4,032 / 500 \\
        Objects-Comparison & 2 & 5,000 / 500 \\
        Antonym-Negation & 2 & 4,779 / 500 \\
        Property-Conjunction & 3 & 4,000 / 483 \\
        Taxonomy-Conjunction & 3 & 5,310 / 599 \\
        Encyclopedic-Composition & 3 & 5,317 / 500 \\
        Multi-Hop Composition & 3 & 5,000 / 500 \\
        \midrule \bottomrule 
    \end{tabular}}
    \caption{The sizes of some probing datasets with fixed number of classes.}
    \label{tab:probing_dataset_sizes_fixed_class}
\end{table}

\begin{table}[t]
    \centering
    \resizebox{.8\linewidth}{!}{
    \begin{tabular}{l l}
        \toprule 
        Suite and Task & N. samples \\ \midrule 
        \multicolumn{2}{l}{LAMA \citep{petroni-etal-2019-language}} \\
        Google-RE / birth-place & 1,937 \\
        Google-RE / birth-date & 1,825 \\
        Google-RE / death-place & 765 \\
        T-REx / 1-1 & 5,527 \\
        T-REx / N-1 & 20,006 \\
        T-REx / N-M & 13,096 \\
        ConceptNet  & 11,458 \\
        SQuAD  & 305 \\ \midrule 
        \multicolumn{2}{l}{CAT \citep{zhou2020evaluating}} \\
        Conjunction Accessibility & 183 \\
        Winograd Schema Challenge  & 283 \\
        Sense Making   & 1,877 \\
        Sense Making with Reasoning  & 2,021 \\
        SWAG  & 1,001 \\
        HellaSWAG  & 1,000 \\
        ability / arct\_1 & 444 \\
        ability / arct\_2 & 888 \\
        \midrule \bottomrule 
    \end{tabular}}
    \caption{The sizes of some probing datasets with no fixed number of classes.}
    \label{tab:probing_dataset_sizes_unlimited_class}
\end{table}
\end{document}